\documentclass{article}
\usepackage{spconf,amsmath,graphicx,hyperref, multirow}
\usepackage{xcolor}

\title{PathFinder: MCTS and LLM Feedback-based Path Selection \\ for Multi-Hop Question Answering}
\name{Durga Prasad Maram$^{1\dagger}$, Kalpa Gunaratna$^{2}$ , Vijay Srinivasan$^{2}$, Haris Jeelani$^{2}$, Srinivas Chappidi $^{2}$\thanks{ $^{\dagger}$ This work was done when Durga Prasad Maram interned at Samsung Research
America.}}

\address{$^{1}$University of Massachusetts Amherst , $^{2}$Samsung Research America \\
dmaram@umass.edu, \{k.gunaratna, v.srinivasan, h.jeelani, vasu.c\}@samsung.com}

\newcommand{\sysname}{{PATHFINDER}}

\begin{document}
\ninept
\maketitle
\begin{abstract}
Multi-hop question answering is a challenging task in which language models must reason over multiple steps to reach the correct answer. With the help of Large Language Models and their reasoning capabilities, existing systems are able to think and decompose an input question over multiple steps to analyze, retrieve, and reason. However, training-based approaches for this problem still suffer from LLM hallucinations and incorrect reasoning paths that hinder performance. Hence, we propose \sysname{}, an approach that: (i) uses Monte Carlo Tree Search to generate training path traces, (ii) improves training data quality by filtering erroneous and lengthy traces using sub-answer recall and LLM-as-a-judge verification, and (iii) reformulates sub-queries to handle failed retrieval cases. By following these steps, we demonstrate that \sysname{} improves the performance of multi-hop QA over public benchmark datasets.  
\end{abstract}
\begin{keywords}
multi-hop question answering, retrieval augmented generation, reasoning, large language models
\end{keywords}
\section{Introduction}
\label{sec:intro}

Large Language Models (LLMs) have demonstrated remarkable capabilities in reasoning-intensive tasks. However, LLMs tend to hallucinate when using only their internal parametric knowledge in answering complex questions. With the advancements of Retrieval Augmented Generation (RAG) \cite{rag,ragsurvey}, LLMs are utilized effectively to answer complex multi-hop questions over unstructured text. Approaches tackling this problem focus on decomposing the input question into multiple sub-questions and finding sub-answers through reasoning. In doing so, they avoid needing to retrieve all facts in the input question at once, which could lead to errors due to noisy or irrelevant context.

In iterative retrieval-based methods\cite{ircot,retplangen,iterdrag,deeprag} for complex multi-hop QA that involves multiple steps of reasoning, question decomposition becomes crucial as: (i) retrieval performance directly depends on decomposed questions, and (ii) sub-answers for these sub-questions decide the ability to reach the final answer. Existing works such as \cite{ircot, atomr, react, searcho1} rely on few-shot prompting to elicit this decomposition, which requires careful handcrafting of Chain-of-Thought (CoT) behavior, large LMs, and longer input context lengths. Thus, training smaller LMs to learn reasoning for iterative retrieval is important for improving efficiency (e.g., deployment in resource and time constrained settings). Recent works such as \cite{selfrag,autorag, deeprag} introduce mechanisms for automatic data synthesis to generate CoT train traces. They dynamically decide when and what to retrieve in the reasoning process while also leveraging LLM’s parametric knowledge. More recent works \cite{searchr1,deepresearcher} leverage reinforcement learning to train LLMs to reason with interleaved search queries dynamically.

However, efforts to validate the correctness and faithfulness of the generated CoT training traces are minimal, as only the final answer correctness and the presence of golden sub-answers are typically evaluated. DeepRAG~\cite{deeprag} further assumes the availability of relevant context by using ground-truth supporting knowledge paragraphs in its data generation process, creating an artificial retrieval capability. As a consequence, this leads to incorrect reasoning in cases of failed retrieval during inference. Moreover, reliance on the parametric knowledge of LLMs can be misleading, sometimes producing an incorrect reasoning even when the final answer is correct. Our evaluation demonstrates that even certain baselines, such as DeepRAG, perform worse when they rely on the LLM's parametric knowledge.

To address these limitations, we propose \sysname{} that: (i) employs Monte Carlo Tree Search (MCTS) to generate diverse CoT traces as training data, (ii) applies a series of trace filtering steps that includes retaining shortest CoT traces, keeping traces grounded in golden sub-answers, and performing LLM-as-a-judge verifications, and (iii) trains the LLM to reformulate sub-questions when sub-answers cannot be found from the retrieved context (reduces incorrect/hallucinated answers). MCTS-based diverse CoT generation and filtering steps enable high-quality traces that lead from input question to final answer. We show that \sysname{} outperforms comparable baselines on public benchmark datasets.

\begin{figure*}[t]
\includegraphics[width=1\textwidth]{Design_Final.pdf}
\caption{Design pipeline of \sysname{}. It involves (i) CoT trace generation with MCTS, (ii) filtering CoT traces with sub-answer verification, (iii) filtering CoT traces with LLM-as-a-Judge, and (iv) training the target LLM with optimal traces based on the LLM Judge scores and the trace length. Intermediate OST and few context steps are omitted above. The LLM Judge validation example demonstrates one of the 4 criteria we consider, the faithfulness of sub-answer generation.}
\label{fig:design}
\end{figure*}

\section{Related work}
\label{sec:related}
 Teaching LLMs to perform iterative retrieval reasoning is both useful and efficient, but obtaining training trajectories at scale is challenging. To address this, AutoRAG\cite{autorag} introduces a mechanism for data synthesis and self-trains the model to develop reasoning with iterative retrieval capabilities. However, it suffers from excessive retrievals and ineffective question decomposition. DeepRAG\cite{deeprag} leverages a larger open-source model to generate training trajectories and calibrates the LLM’s internal knowledge to decide when to retrieve external information. %
Recent works have explored the use of tree-search methods, such as MCTS, for solving complex multi-hop questions. MCTS-RAG \cite{mctsrag} and Air-RAG \cite{airrag} rely on inference scaling, where multiple reasoning paths are explored to increase the likelihood of finding the solution, with the optimal answer chosen via voting or semantic similarity. Although MCTS at inference time boosts performance, it comes at the cost of increased latency and computational complexity. In comparison, our proposed approach relies on fine-tuned models and hence these systems are orthogonal to ours. However, MCTS can serve as a powerful tool for data generation, given its ability to diversify the search space, provide multiple correct paths, and thereby increase the likelihood of identifying the most optimal solution. RaDeR \cite{rader} leverages MCTS to generate synthetic data for reasoning-intensive retrieval. Prior works such as \cite{mctsdpo,restmcts} also demonstrate this, but they primarily focus on math and coding benchmarks, while the effectiveness of MCTS-based data generation for better multi-hop question answering remains underexplored.

\section{Approach}
\label{sec:approach}

We describe our approach \sysname{} in detail along answer trace generation, different filtering steps, and training pipeline. Figure~\ref{fig:design} illustrates and describes main components in \sysname{} using some real examples as well.

\subsection{Answer Trace Generation}
The data generation process using MCTS is driven by the action space and reward propagation. Inspired by ReACT\cite{react}, we adopt a simplified action space consisting of subquestion generation and thought generation. More specifically:
\\
\textbf{$A1$: One-Step Thought (OST):}
This action generates an immediate thought based on the current reasoning chain. It analyzes what information is next needed to answer the original question and help formulate the next subquestion. This action can also represent a terminal state if it outputs the final answer marker.
\\
\textbf{$A2$: Subquestion Generation and Answering (SGA):}
This action formulates a retrieval query based on the current reasoning chain. A retrieval step is performed and based on the retrieved knowledge, an answer is generated and added to the reasoning chain.

Current reasoning chain is the path from root to the current node. $A1$ and $A2$ alternate sequentially, and the chain is terminated when the final answer is given in $A1$ or when the maximum depth of the tree is reached. The input question is followed by an $A1$ step. Importantly, we use the sampling temperature to generate multiple instances of an action at a particular step, thereby enabling branching at the corresponding node of the tree. Specifically, we set the number of sampling nodes to 2 for $A1$, and 3 for $A2$. We use few-shot prompting for both actions.

\begin{table*}[]
\centering
\small
\setlength{\tabcolsep}{3pt}
\begin{tabular}{lccccccccccccc}
\hline
\multicolumn{1}{c}{\multirow{3}{*}{Method}} & \multicolumn{6}{c}{in-distribution}                     & \multicolumn{1}{l}{} & \multicolumn{6}{c}{out-of-distribution}                         \\ \cline{2-7} \cline{9-14} 
\multicolumn{1}{c}{}                        & \multicolumn{3}{c}{2WikiMultiHopQA} & \multicolumn{3}{c}{MuSiQue} & \multicolumn{1}{l}{} & \multicolumn{3}{c}{HotpotQA} & \multicolumn{3}{c}{WebQuestions} \\ \cline{2-7} \cline{9-14} 
\multicolumn{1}{c}{}                        & EM     & Accuracy & F1    & EM     & Accuracy  & F1     & \multicolumn{1}{l}{} & EM      & Accuracy  & F1     & EM       & Accuracy    & F1      \\ \hline
\multicolumn{14}{c}{Qwen-2.5-7B-Instruct}                                                                                                                                                      \\ \hline
DeepRAG (Imi)                               & 47.70  & 48.90    & 52.87 & 18.40  & 20.80     & 25.37  &                      & 27.80   & 29.50     & 37.72  & 25.30    & 28.40       & 38.01   \\
DeepRAG (Imi-retrieve only)                 & 69.40  & 70.90    & 73.48 & \textbf{32.50}  & 35.00     & 40.53  &                      & 39.10   & 41.70     & 50.81  & \textbf{29.30}    & 33.80       & \textbf{42.49}   \\ \hline
Pathfinder                                  & \textbf{75.00 } & \textbf{76.30}    & \textbf{78.38} & 32.10  & \textbf{35.20}     & \textbf{40.86}  &                      & \textbf{39.70}   & \textbf{41.90}     & \textbf{51.88}  & 28.50    & \textbf{38.10}       & 42.47   \\ \hline
\multicolumn{14}{c}{Gemma-2-9B-Instruct}                                                                                                                                                       \\ \hline
DeepRAG (Imi)                               & 58.00  & 59.70    & 62.06 & 28.30  & 31.10     & 36.22  &                      & 32.80   & 34.90     & 44.00  & 23.60    & 27.70       & 36.18   \\
DeepRAG (Imi-retrieve only)                  & 71.90  & 73.80    & 75.46 & 31.40  & 34.30     & 40.68  &                      & 36.40   & 39.00     & 47.56  & 29.30    & \textbf{34.20}       & 41.92   \\ \hline
Pathfinder                                  & \textbf{72.50}  & \textbf{74.60}    & \textbf{76.77} & \textbf{33.50}  & \textbf{38.10}     & \textbf{42.41}  &                      & \textbf{36.60}   & \textbf{39.40}     & \textbf{47.91}  & \textbf{30.60}    & 34.10       & \textbf{44.08}   \\ \hline
\end{tabular}

\caption{Main evaluation results using Qwen and Gemma as finetuned LLMs using in-distribution and out-of-distribution datasets, where in-distribution datasets are the ones used to finetune the respective LLMs. Training data generation used Qwen 72B Instruct for all approaches.}
\label{tab:main_result}
\end{table*}

\paragraph*{MCTS Traces} During each iteration of the MCTS process, we perform the selection, expansion, simulation, and backpropagation steps. To balance exploitation and exploration, we use Upper Confidence Bounds applied to trees (UCT) scores during the selection step. 

\[
UCT(s, a) = \frac{Q(s, a)}{N(s,a)} + w \sqrt{\frac{\log N(s)}{N(s,a)}}
\]
where $Q(s,a)$ is the cumulative reward for node $s$ obtained by taking action $a$, and is updated through reward backpropagation, and $N(s)$ is the number of visits to node $s$. $w$ is the weight balancing exploration and exploitation.
Starting from the root node, the child node with the highest UCT value is chosen at each level until a leaf node is reached. If the leaf node is not terminal, it is expanded by adding all possible child nodes. A random node is then selected from the newly added child nodes, and the simulation step is performed by executing the next action and randomly selecting a child node until a terminal node or the maximum depth is reached. Finally, a reward is calculated at the terminal node and back propagated along the path back to the root. A terminal node here is an OST ($A1$) step that contains a final answer marker. The reward at the terminal node is decided by the correctness of the predicted answer: 1 if the gold answer is present in the prediction, and 0 otherwise. 
All the tree paths that lead to the correct answer are potential CoT train traces.

\subsection{Filtering Answer Traces}

After running MCTS, each question may have multiple correct solution traces. To select the most optimal trace from this collection, we follow a two-stage filtering process. 

In the first stage, we verify the recall of golden sub-answers in the generated outputs of the $A2$ step. All traces with a sub-answer recall not equal to 1 are discarded. This ensures that the reasoning process in the CoT trace is factually correct and that all relevant information necessary for answering the question is retrieved. See the top right of Figure~\ref{fig:design}.

In the second stage, we leverage LLM-as-a-Judge to evaluate the quality and faithfulness of a CoT trace. See the bottom right of Figure~\ref{fig:design}. We validate the quality of each solution trace along 4 different aspects. 
\\
\textbf{(i) Incorrectness: } Evaluate whether the reasoning process in the trace is logically consistent and whether each step remains strictly grounded in the preceding steps. Errors are identified when a sub-question or a thought step introduces unsupported entities or information (not present in previous steps), makes incorrect inferences, contradicts prior reasoning, or draws unwarranted conclusions.
\\
\textbf{(ii) Redundancy: } Assess whether sub-questions are redundant within the reasoning process. Redundancy errors occur when a sub-question attempts to retrieve information that has already been successfully obtained earlier, although reformulations following a failed retrieval attempt are acceptable.
\\
\textbf{(iii) Irrelevance: } Examine whether sub-questions are unnecessary and do not contribute to reaching the final answer. Such errors arise when the answer to a sub-question is neither used directly in obtaining the final answer nor leveraged to reach another sub-question that is useful.
\\
\textbf{(iv) Sub-Answer Generation Faithfulness: } For each sub-question, we evaluate whether the model’s generated answer is faithful to the sub-question and supported by the retrieved context. Errors occur when: the answer lacks clear supporting evidence in the context, the model fails to provide an answer despite sufficient evidence, the context itself is insufficient but the model does not state ‘not found’ or something similar, or the answer is irrelevant to the question.

The LLM Judge is instructed to output the number of erroneous steps in the CoT, along with a justification for each. If the LLM Judge flags issues related to the incorrectness or faithfulness, those traces are discarded. For the remaining traces, a weighted error score is assigned based on the number of redundant and irrelevant error steps.
When choosing the final optimal trace for a question, we choose the trace with the minimum error score from the filtered trace pool. In case of a tie, we pick the one with the shortest length.

\subsection{Training}
Once we obtain the final training set with the optimal traces for all questions, we fine-tune the target LLM on the causal language modeling objective. The context documents are masked, and no loss is computed over them. The LLM effectively learns to generate relevant thoughts and sub-queries, as well as intermediate answers, based on the retrieved context. The training objective is as follows:
\[
\begin{aligned}
L = &- (\sum_{i=1}^{n-1} (\log\Pr(t_i,q_i \mid c_{<i}) + \log\Pr(a_i \mid c_{<i}, t_i, q_i, d_i))  \\ &+ \log\Pr(t_n \mid c_{<n}))
\end{aligned}
\]
where $c_{<i}$ is the entire reasoning chain before step i which consists of $\{ t_{<i}, q_{<i}, d_{<i}, a_{<i} \}$. $q_i$ is the sub-query, $d_i$ is the retrieved context for $q_i$, $a_i$ is the sub-answer, and $t_i$ is the thought at the $i^\text{th}$ step.
The final step, corresponding to the final OST $t_n$, contains the final answer marker as well as the correct answer. $n$ is the trace length.

\section{Evaluation}
\label{sec:evaluation}

We evaluate \sysname{} on four QA datasets. 2WikiMultiHopQA \cite{2wiki} and MuSiQue\cite{musique} are evaluated in the in-distribution setting, as they are used to generate the training data, while the out-of-distribution datasets consist of HotpotQA \cite{hotpot} and WebQuestions~\cite{webquestions}. Following \cite{deeprag}, we choose the first 1000 samples from the dev splits of each dataset for our evaluation. We consider DeepRAG \cite{deeprag} as our baseline as it is the highest-ranked comparable method reported in the literature. We reproduce DeepRAG's imitation learning configuration in its original setting (Imi), which uses parametric knowledge, as well as in a retrieval-only setting (Imi-retrieve-only), where external context is used for every follow-up query. We use Qwen \cite{qwen2} and Gemma\cite{gemma2} instruction-tuned LLMs for our evaluation~\footnote{For legal licensing issues that are enforced on certain corporate levels, we cannot report Llama models in the evaluations. Initial experiments revealed even better performance favoring \sysname{}.}. We report Exact Match (EM), Accuracy: which measures if the gold answer is completely included in the prediction, and token-level F1 scores.

\paragraph*{Implementation Details}
For MCTS based data generation, we rollout the tree 12 times with max depth set to 12. The number of child nodes added for action $A1$ is 2, while for action $A2$ is 3. For $A1$, we set the sampling temperature to 0.6. For $A2$, the sub-question generation temperature is set to 1, to increase the chance of generating a retriever aligned sub-query, while the answer extraction temperature is set to 0.2. Qwen-2.5-72B-Instruct is the base model used for training data generation for \sysname{} and DeepRAG.

We use Qwen-2.5-7B-Instruct and Gemma-2-9B-Instruct as our target models for training. For the training dataset, we randomly sampled 4,000 questions from 2WikiMultiHopQA and 4,000 questions from MuSiQue to train \sysname{} and DeepRAG. We use MCTS to generate CoT traces for \sysname{}. From the total 8000 questions, we ended up having 6852 after the filtering steps for \sysname{} finetuning. We employed Qwen-2.5-72B-Instruct as the base model to run MCTS for enhanced reasoning trace generation. ColBERTv2 \cite{colbertv2} is used as the retriever, and we utilize the December 2021 full Wikipedia dump from Atlas \cite{atlas} as our knowledge corpus. We set top-k to 3, appending the top three passages as context for each retrieval query. For LLM-as-Judge, we use the Qwen-2.5-72B-Instruct model. We trained the models on AMD MI300X GPUs. We trained DeepRAG for 2 epochs. ~\sysname{} is trained for 5 epochs and 2 epochs for Qwen and Gemma, respectively.

\begin{figure}[t]
\includegraphics[width=1\columnwidth]{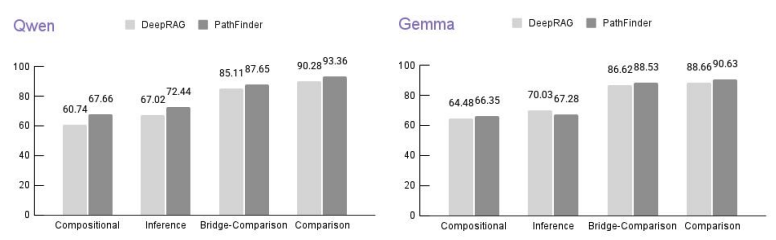}
\caption{DeepRAG (Imi-retrieve-only) vs. \sysname{} on F1 score across different question types in 2WikiMultiHopQA. }
\label{fig:comparison}
\end{figure}

\subsection{Results \& Discussion}

We evaluate \sysname{} in both in-distribution and out-of-distribution settings, as shown in Table ~\ref{tab:main_result}. \sysname{} achieves better results than DeepRAG (Imi-retrieve-only) for a majority of in-domain and out-of-domain datasets across models due to its high-quality answer-trace generation and sub-question reformulation capability. \sysname{} also makes significantly fewer errors than DeepRAG (Imi), as it relies only on retrieved context rather than parametric knowledge, which can cause hallucinations. We observe that DeepRAG performs better without LLM memory in our setting with a strong Colbert retriever, as LLM memory can introduce hallucinations. In contrast, the DeepRAG paper found little difference between memory and retrieve-only modes, likely due to using a weaker BM25 retriever.

We validate the effectiveness of our filtering mechanism as shown in the ablation study in Table~\ref{tab:ablation_study}. We run \sysname{} in three configurations. In the initial configuration, for every question, we consider the shortest CoT trace from the correct solution trace collection as the final train sample. In the second setting, we do the first-stage sub-answer recall verification, and pick the shortest CoT from the filtered pool for each question. In the third and the best setting, after running LLM Judge filtering, we pick the CoT with minimum error score. If there is a tie, we choose the one with shortest length as the final train sample. While each filtering step improves answer accuracy, we see the biggest jump in accuracy with the LLM validation step.

\begin{figure}[!t]
\includegraphics[width=1\columnwidth]{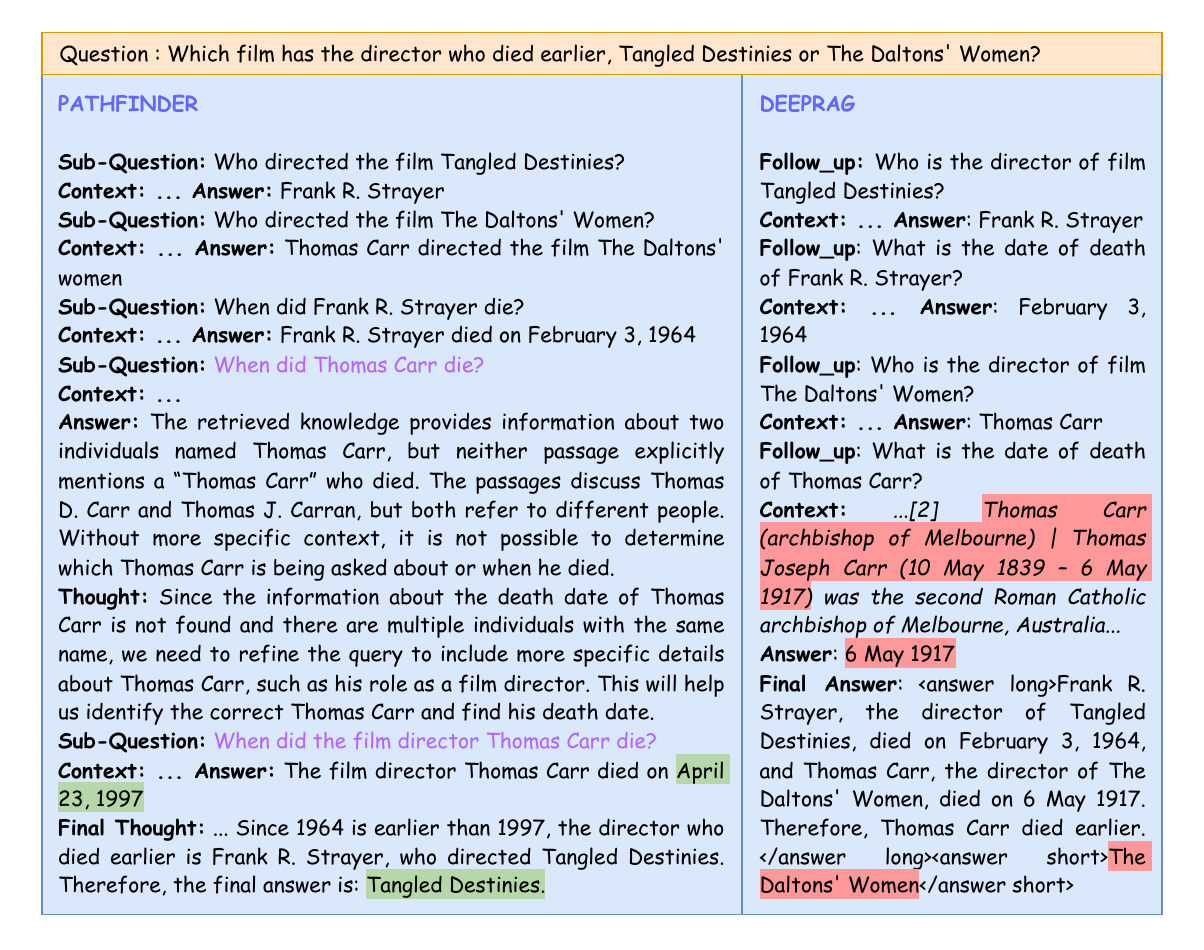}
\caption{\sysname{} vs DeepRAG. \sysname{} finds the right answer through effective query reformulation (highlighted sub-questions in purple) when relevant context is not retrieved.}
\label{fig:inference_example}
\end{figure}

We also analyze and breakdown the performance of our approach \sysname{} against DeepRAG on the complexity of the questions in Figure \ref{fig:comparison}. We can see that \sysname{} performs equally well in all levels of complexity compared to the state-of-the-art baseline DeepRAG. An inference time reasoning trace comparison between our proposed approach ~\sysname{} and DeepRAG\cite{deeprag} (Imi-retrieval only) is illustrated in Figure \ref{fig:inference_example}. When the retrieved context does not answer the question, \sysname{} tries to re-query the retriever by reformulating with more relevant information in the next sub-question to disambiguate the entity/information, instead of producing an irrelevant extracted answer like DeepRAG, which leads to incorrect reasoning.

\begin{table}[htbp]
\centering
\footnotesize
\begin{tabular}{lcrrrccr}
\hline
\multicolumn{1}{c}{\multirow{3}{*}{Configuration}} & \multicolumn{4}{c}{In-Distribution}                                                                        &                      & \multicolumn{2}{c}{OOD}                           \\ \cline{2-5} \cline{7-8} 
\multicolumn{1}{c}{}                               & \multicolumn{2}{c}{2Wiki}                          & \multicolumn{2}{c}{MuSiQue}                     &                      & \multicolumn{2}{c}{HotpotQA}                      \\ \cline{2-8} 
\multicolumn{1}{c}{}                               & EM                        & \multicolumn{1}{c}{F1} & \multicolumn{1}{c}{EM} & \multicolumn{1}{c}{F1} &                      & EM                       & \multicolumn{1}{c}{F1} \\ \hline

SP                                     & 71.40       & 75.56       & 28.00     & 38.62  & \multicolumn{1}{r}{}        & 39.30         & 50.93        \\
SP + AV                                & 71.90       & 75.61       & 29.00        & 39.02  & \multicolumn{1}{r}{}        & 39.20         & 51.03        \\
SP + AV + LJ                           & \textbf{75.00}       & \textbf{78.38}       & \textbf{32.10}        & \textbf{40.86}  & \multicolumn{1}{r}{}        & \textbf{39.70}         & \textbf{51.88}       \\ \hline
\end{tabular}

\caption{Ablation analysis of \sysname{} using Qwen-2.5-7B-Instruct model. SP, AV, and LJ represent shortest path, answer verification, and LLM-as-a-Judge, respectively.}
\label{tab:ablation_study}
\end{table}

In future work, we plan to investigate how to more explicitly align the sub-question generation with the retriever (e.g., using preference learning from retrieval success and failure question pairs generated by MCTS). Moreover, exact match and F1 scores do not always reflect answer correctness, as supplementary words (e.g., ‘London, UK’ vs. ‘London, England’) may unfairly affect scores. Using LLMs to evaluate final answers can partly mitigate this issue.

\section{Conclusion}
\label{sec:conclusion}

We presented \sysname{}, a multi-hop complex question answering framework that trains LLMs using high quality answer traces. It generates diverse gold answer traces using MCTS and performs several filtering steps to ensure better training traces that resulted in improved fine-tuned LLMs for multi-hop QA with RAG. While the use of LLMs' internal parametric knowledge seems a popular choice in this domain, we showed that it is more important to rely on the retrieved context.

\vfill\pagebreak

\bibliographystyle{IEEEbib}
\bibliography{references}

\end{document}